\documentclass[conference]{IEEEtran}
\IEEEoverridecommandlockouts

\usepackage{cite,url}

\ifCLASSINFOpdf
   \usepackage[pdftex]{graphicx}
\else
\fi

\usepackage[cmex10]{amsmath}

\usepackage[tight,footnotesize]{subfigure}

\begin{document}

%
\title{Persian Heritage Image Binarization Competition (PHIBC 2012)}

\author{\IEEEauthorblockN{Seyed~Morteza~Ayatollahi~and~Hossein~Ziaei~Nafchi}
\IEEEauthorblockA{Synchromedia Laboratory for Multimedia Communication in Telepresence,\\
\'Ecole de technologie sup\'erieure, Montreal (QC), Canada H3C 1K3\\
Tel.:  +1(514)396-8972\\
Fax: +1(514)396-8595\\
Email: sr.morteza@gmail.com, hossein\_zi@yahoo.com}\\
}

\maketitle


\begin{abstract}


The first competition on the binarization of historical Persian documents and manuscripts (PHIBC 2012) has been organized in conjunction with the first Iranian conference on pattern recognition and image analysis (PRIA 2013). The main objective of PHIBC 2012 is to evaluate performance of the binarization methodologies, when applied on the Persian heritage images. This paper provides a report on the methodology and performance of the three submitted algorithms based on evaluation measures has been used.  \IEEEpubidadjcol

\end{abstract}


\IEEEpeerreviewmaketitle

\section{Introduction}
There are many old manuscripts and documents in the libraries and museums of Iran. Many of them include historically important data which needs automatic processing and reading. However, less attention has been made to preserve these valuable types of documents. The PHIBC 2012 is a primitive attempt toward evaluation of binarization methods, when applied on the Persian manuscripts.

Persian heritage documents and manuscripts following their similarity to Arabic documents \cite{sina} are in the form of handwritten document images. As usual, for handwritten documents, preservation of strokes and sub-strokes is of great interest. Previously, five Latin datasets of historical manuscripts has been publicly available for DIAR researchers \cite{D2009,D2010,D2011,D2012,bleed2012}. PHIBC 2012 introduces the first dataset that developed for binarization of Persian heritage documents. The dataset used for PHIBC 2012 consisted of ten historical documents used with permission from "Documents and old manuscripts treasury of Mirza Mohammad Kazemaini (affiliated with Hazrate Emamzadeh Jafar), Yazd, Iran". The images in PHIBC 2012 suffered from various types of degradation, including bleed-through, deterioration of cellulose structure, faded ink and alien ink, among others \cite{special}. For each image in the dataset, a ground truth is generated with a semi-automatic approach. At first, document is processed with phase congruency features used in \cite{ziaeiaccv,ziaeiicisp} to produce a rough binarized image. Then, the final ground truth is generated manually by human expert from the rough binary image produced in the first step. We will report the ground truth generation methodology in a dedicated report. Figure 1 shows sample images and corresponding ground truth images used in the PHIBC 2012.

\begin{figure*}[tb]
\centering
\begin{tabular}{ccc}
\fbox{{\includegraphics[height = 4cm]{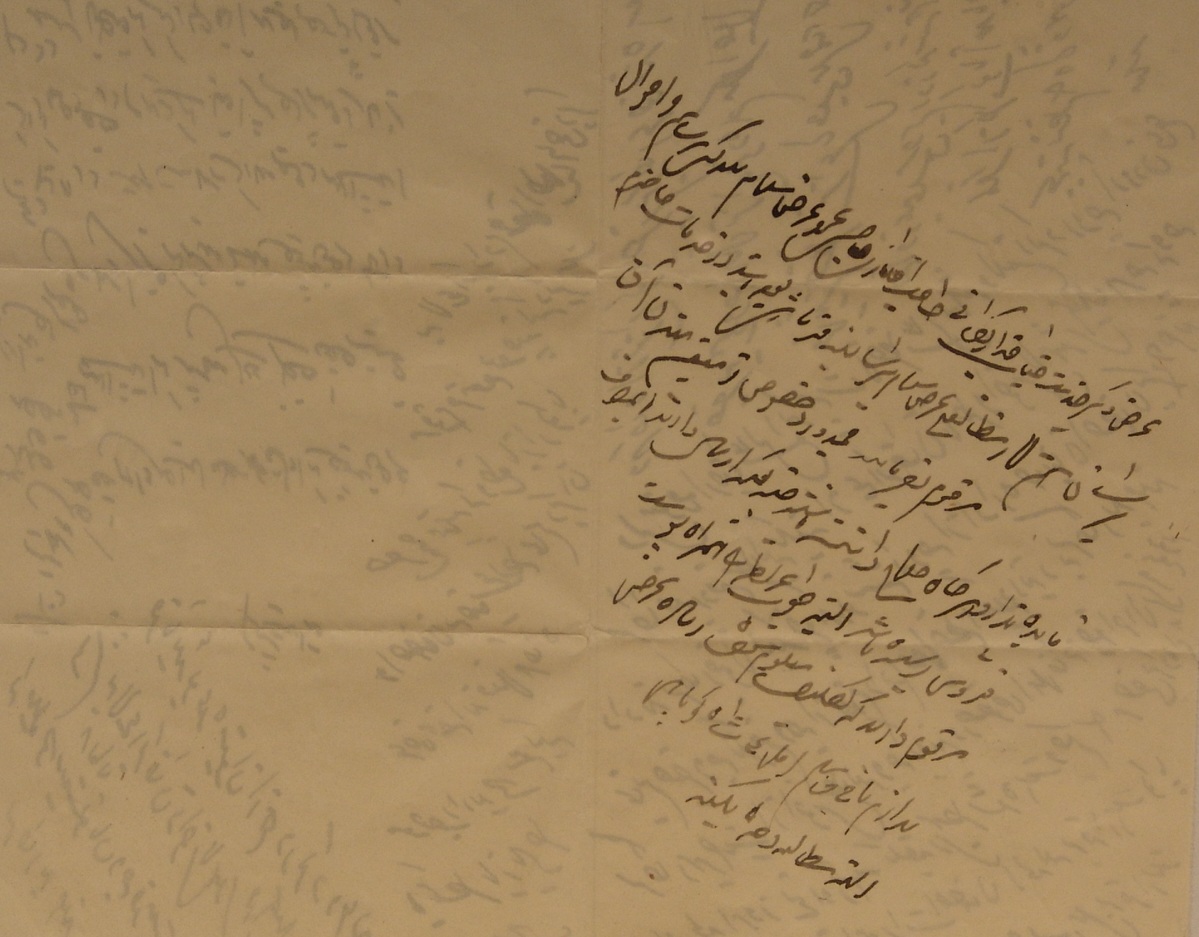}}} &
\fbox{{\includegraphics[height = 4cm]{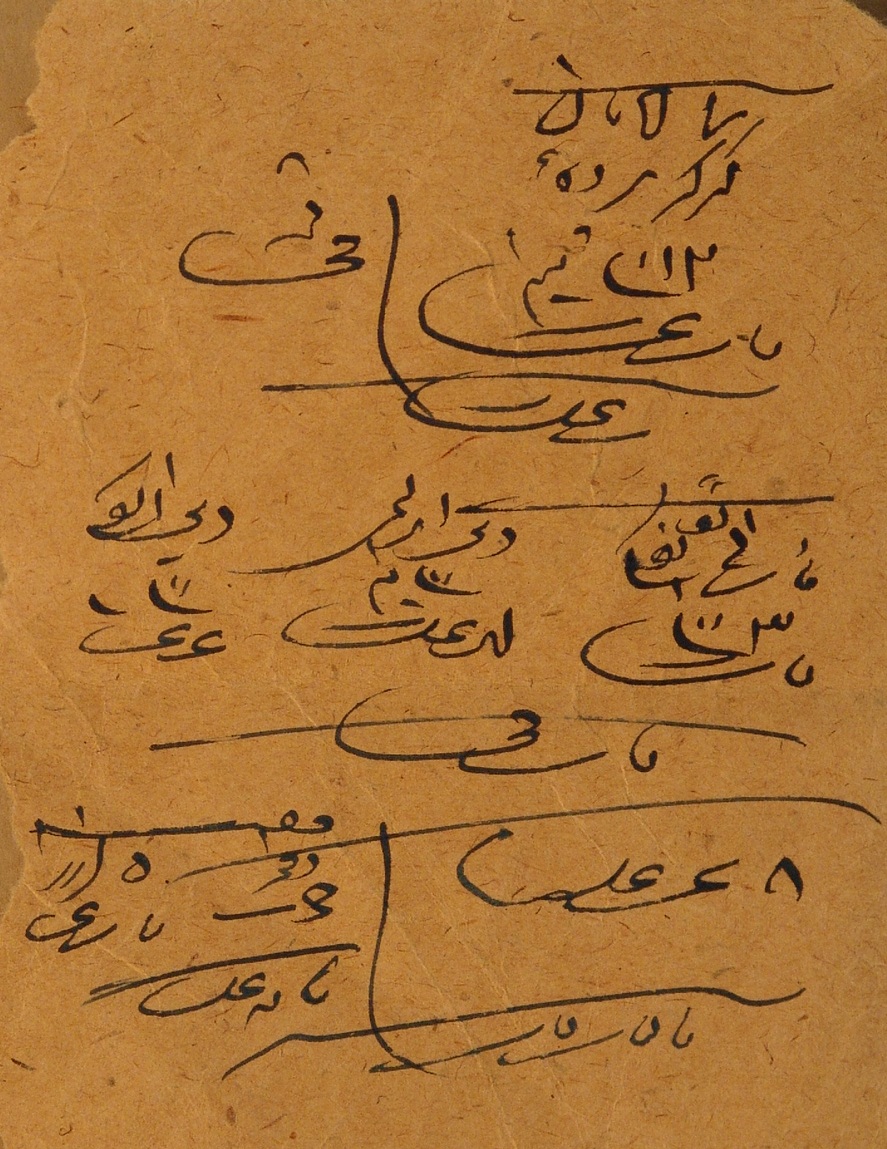}}} &
\fbox{{\includegraphics[height = 4cm]{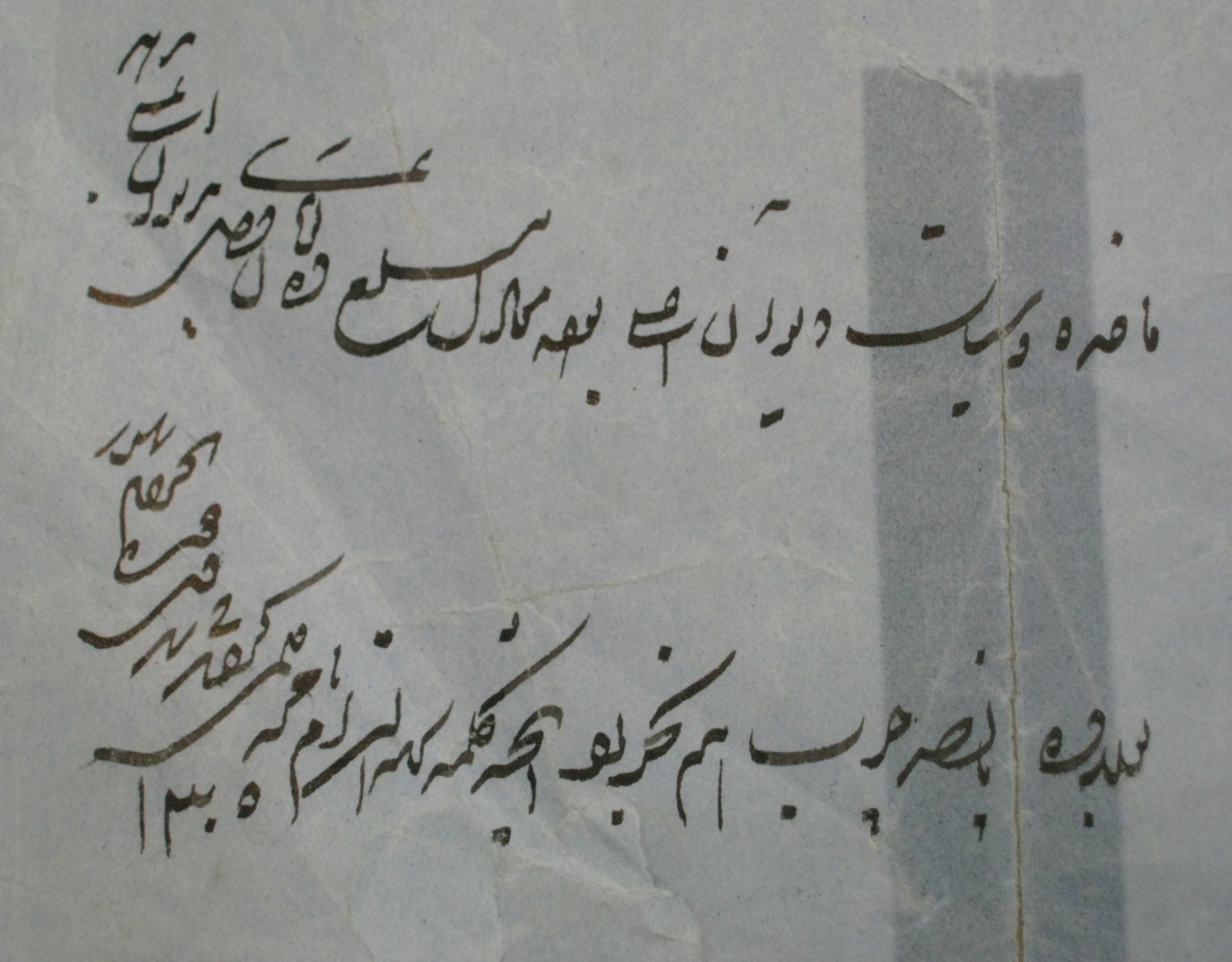}}}
 
\\\\
\fbox{{\includegraphics[height = 4cm]{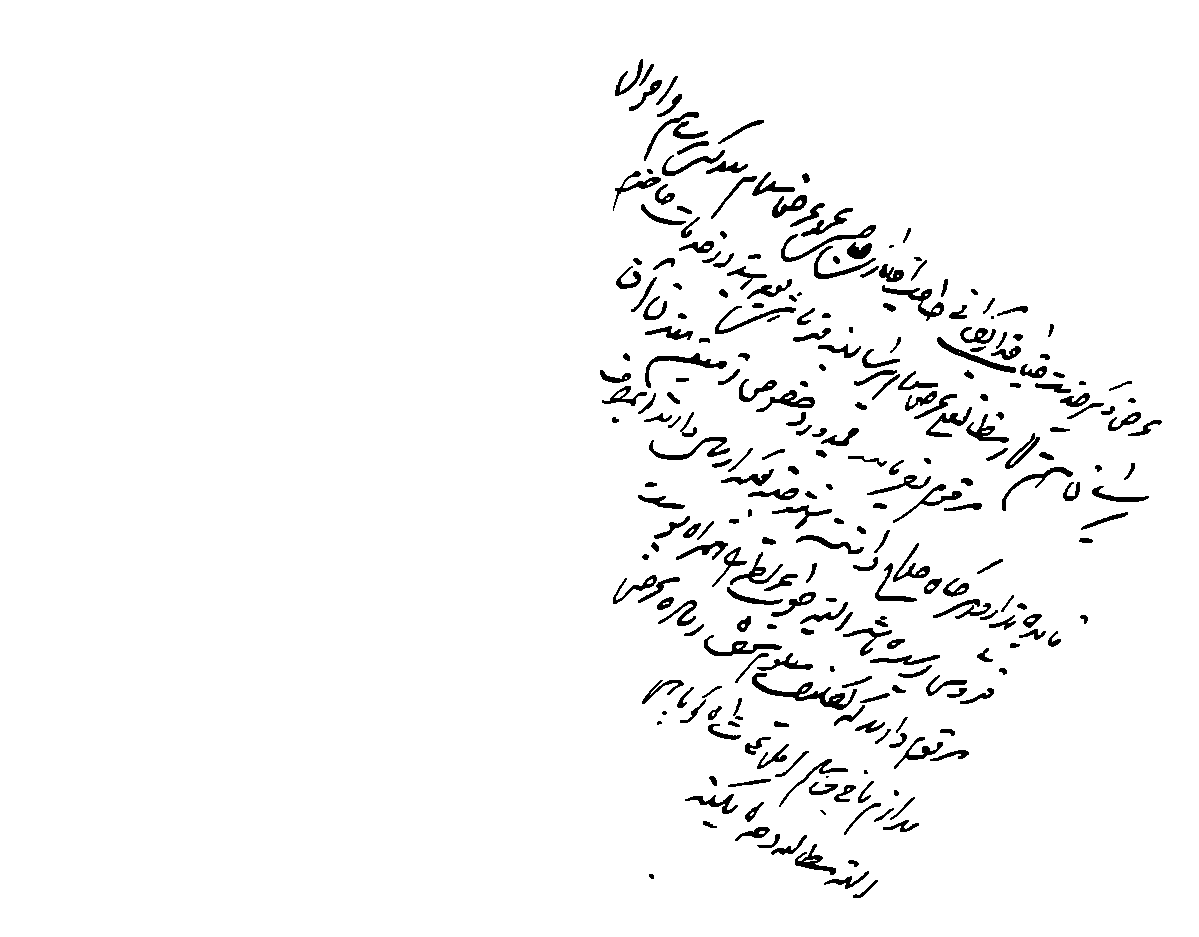}}} &
\fbox{{\includegraphics[height = 4cm]{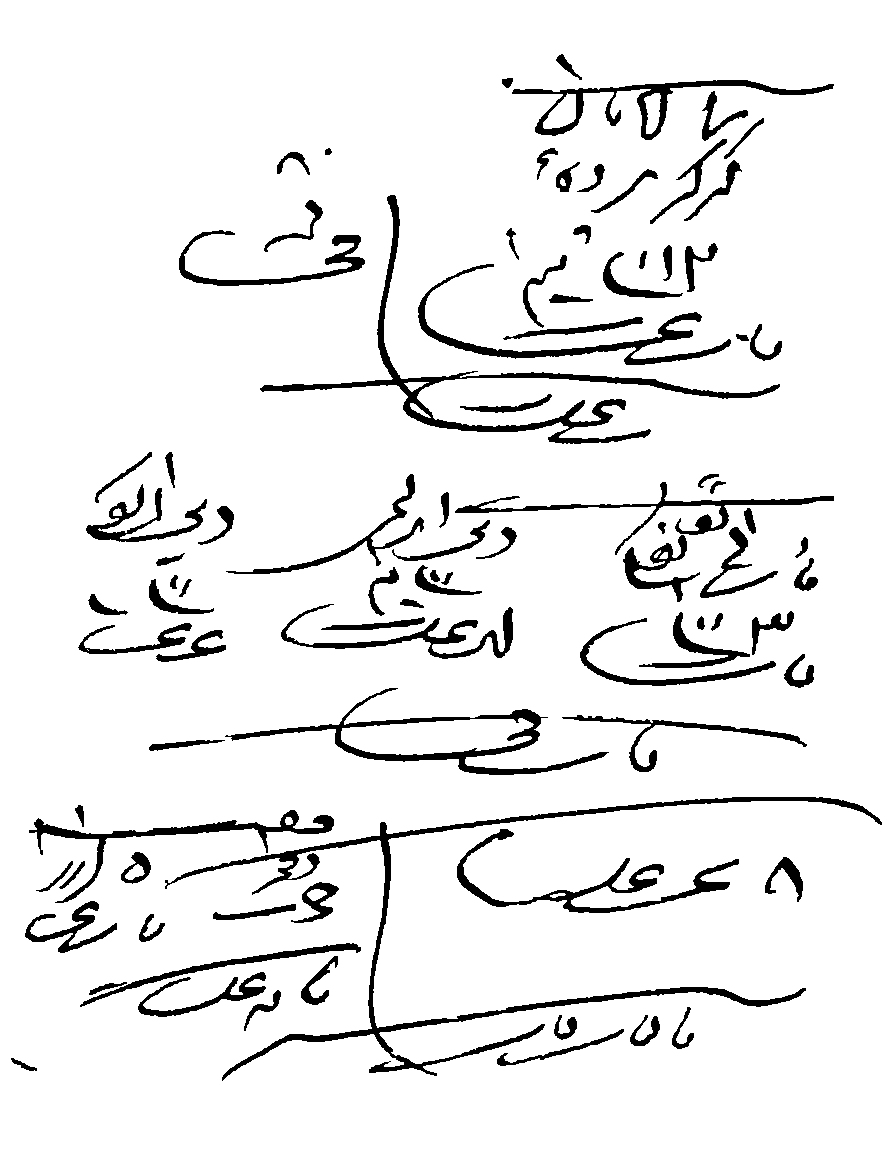}}} &
\fbox{{\includegraphics[height = 4cm]{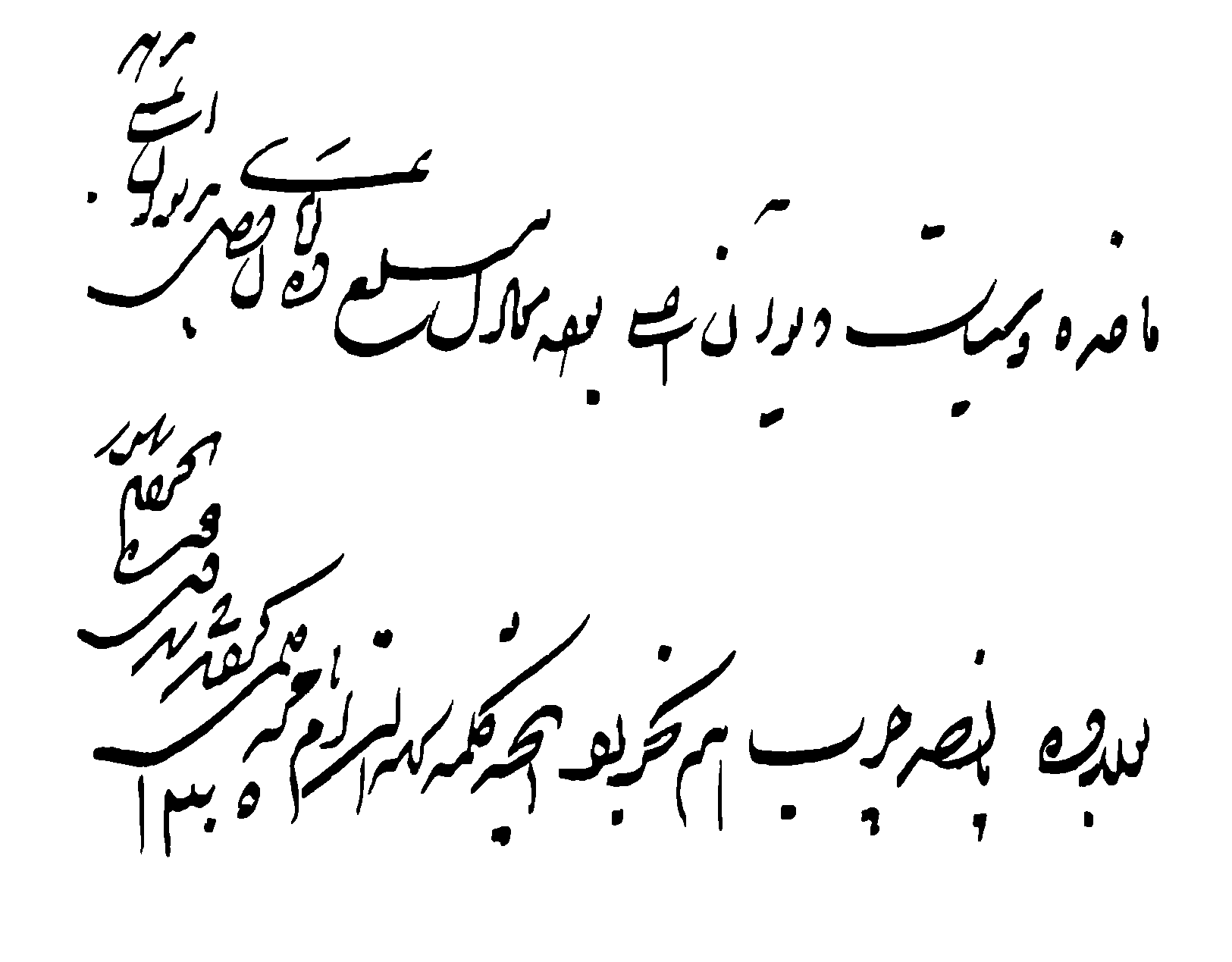}}}

\end{tabular}
\caption {Sample original and ground truth images used in PHIBC 2012.}
\label{f-outputs}
\end{figure*}

The dataset is available via competition website and technical committee 11 website (IAPR TC-11):\\
\url{http://phibc2012.ir/}\\
\url{http://www.iapr-tc11.org/}

Also, the source code of the evaluation measures used for performance evaluation can be found at \cite{math}. 

The rest of the paper is organized as follows. In section II, a brief description of the submitted methods to PHIBC 2012 is provided. The evaluation measures used for comparison between submitted algorithms are described in section III. Section IV provides experimental results. Finally, section V draws a conclusion. 


\section{Description of methodologies}

In the Persian heritage image binarization competition (PHIBC 2012), three groups submitted three algorithms. The description of each methodology is provided by these groups and is as follows.\\\\
\textbf {1-} Su Bolan\(^\dagger\), Tian Shangxuan\(^\dagger\), Lu Shijian\(^\ddagger\) and Tan Chew Lim\(^\dagger\) (\(^\dagger\)School of Computing, National University of Singapore, and \(^\ddagger\)Department of Computer Vision and Image Understanding Institute for Infocomm Research, Singapore).

There are four main steps in our proposed method. First, local image contrast which is evaluated by local maximum and minimum and local image gradient are combined using an exponential function with an adaptive factor. 
Second, The local image contrast is combined with the edge map to extract an accurate text character edge image. Third, the document image is binarized by a local threshold which is decided based on the constructed edge map and estimated stroke width. At last, some post-processing work is applied to produce better results.\\\\
\textbf {2-} Syed Ahsen Raza Ali Hamdani (National University of Sciences and Technology (NUST), Islamabad, Pakistan).

The algorithm is based on three processing steps: preprocessing, thresholding and postprocessing. In preprocessing, conditional noise removal and edge based processing is performed. Thresholding step involves a computation of final threshold for background and text segmentation based on an average value computed through multiple thresholds (based on 4 different Niblack inspired thresholding formulas). In final step of post processing, again conditional noise removal and constrained morphological operations are performed to get the final binarised image.\\\\
\textbf {3-}\footnote {Hereinafter, we refer to each group with its assigned number.} Seyed Mehrdad Kankanan and Hossain Poyarad (Faculty of engineering, Shahid Chamran University of Ahvaz, Ahvaz, Iran).

Proposed method is mainly based on fuzzy measures introduced in \cite{fuzzy2010}. It finds a global threshold and apply it for whole image. 
The main advantage of this method against Otsu's method, is its better classification of bleed-through in the case of documents that 
include large amount of this type of degradation. Afterward, an approach similar to Niblack and some post-processing processes are applied to improve the final binarization result.

\section{Evaluation measures}

For an objective comparison between submitted algorithms, six evaluation measures used \cite{math}. These are F-Measure \cite{measures}, pseudo F-Measure \cite{gatos2008}, peak signal-to-noise ratio (PSNR), distance reciprocal distortion metric (DRD) \cite{DRD}, misclassification penalty metric (MPM) \cite{MPM} and negative rate metric (NRM) \cite{D2009,D2010}. 

Let's \(bwout\) and \(GT\) denotes the binarized image and ground truth image, respectively. PSNR for binary images can be defined as:

\begin{equation}
\text{MSE}=\frac{\sum_{x=1}^{N}\sum_{y=1}^{M} [GT(x,y)-bwout(x,y)]}{N \times M} ~
\end{equation}

\begin{equation}
\text{PSNR}=10 \times \log (\frac{1}{\text{MSE}}) ~.
\end{equation}

F-Measure can be considered as an intelligent alternative for PSNR because it takes into account the number of foreground and background pixels. Let's \(TP\), \(FP\), \(FN\) and \(TN\) denote the true positive, false positive, false negative and true negative, respectively. Recall, precision and F-Measure can be defined as:

\begin{equation}
\text{Recall}=\frac{TP}{TP+FN} ~
\end{equation}

\begin{equation}
\text{Precision}=\frac{TP}{TP+FP} ~
\end{equation}

\begin{equation}
\text{F-Measure}=\frac{2 \times \text{Recall} \times \text{Precision}}{\text{Recall}+\text{Precision}} ~.
\end{equation}

Also, pseudo F-Measure is computed like F-Measure except that recall value is taken from skeletonized ground truth:

\begin{equation}
\text{pseudo~F-Measure}=\frac{2 \times \text{Recall}_{skel} \times \text{Precision}}{\text{Recall}_{skel}+\text{Precision}} ~.
\end{equation}

Furthermore, NRM can be computed as:

\begin{equation}
\text{NRM}=\frac{NR_{FN}+NR_{FP}}{2} ~.
\end{equation}
where:
\begin{equation}
NR_{FN}=\frac{FN}{FN+FP}~~,~~NR_{FP}=\frac{FP}{FP+TN} ~
\end{equation}

DRD measures the distortion for all the S flipped pixels as follows:

\begin{equation}
\text{DRD}=\frac{\sum_{k=1}^{S}\text{DRD}_k}{\text{NUBN}} ~.
\end{equation}

where, \(\text{DRD}_k\) is the distortion of the \(k-th\) flipped pixel and it is calculated using a \(5\times5\) normalized weight matrix \(W_{Nm}\) \cite{DRD}. \(\text{DRD}_k\) equals to the weighted sum of the pixels in
the \(5\times5\) block of the ground truth GT that differ from the centered \(k-th\) flipped pixel at \((x,y)\) in the binarization result image B.

\begin{equation}
\text{DRD}_k=\sum_{i=-2}^{2}\sum_{j=-2}^{2} [GT_k(i,j)-bwout_k(x,y)]\times W_{Nm}(i,j)
\end{equation}

NUBN is the number of the non-uniform (not all black not all white pixels) \(8 \times 8 \) blocks in the \(GT\) image. MPM is a measure of how well the resulting image representing the contour of ground truth image and defined as:

\begin{equation}
\text{MPM}=\frac{1}{2D}(\sum_{i=1}^{FN}d^i_{FN}+ \sum_{j=1}^{FP}d^j_{FP})~.
\end{equation}

where, \(d^i_{FN}\) and \(d^j_{FP}\) denote the distance of the \(i-th\) false negative and the \(j-th\) false positive pixel from the contour of the text in the ground truth image. The factor \(D\) is the sum of all the pixel-to-contour distances of the ground truth object.

A higher value for F-Measure, pseudo F-Measure and PSNR measures, indicate to better classification, while a lower value for DRD, MPM and NRM measures, shows better performance.

\section{Experimental results}

In this section, characteristics of test images used in the PHIBC 2012 are described. Images in the PHIBD 2012 suffered from various types of degradation, include uneven illumination changes, various types of bleed-through, etc. Table I has summarized the degradation types of each document image used in the PHIBC 2012.

\setlength{\tabcolsep}{4pt}
\begin{table*}[!ht]
\begin{center}
\caption{
Characteristics of test images used in the PHIBC 2012
}
\label{table2}
\begin{tabular}{lll}
\hline\noalign{\smallskip}
 Image name & Size & Degradation type(s) \\
\noalign{\smallskip}
\hline
\noalign{\smallskip}
Persian01 & {\ 1625\(\times\)1269} ~~~& faded ink, multi-degraded background, color background.\\
Persian02   & {\ 845\(\times\)691} ~~~&  bleed-through, alien ink, low resolution. \\
Persian03  & {\ 1215\(\times\)735} ~~~& bleed-through, degraded background, small amount of text.\\
Persian04  & {\ 1247\(\times\)1829} ~~~& faded ink, color background, degraded background. \\
Persian05  & {\ 887\(\times\)1149} ~~~& faded ink, color background, visible fibers in the paper. \\
Persian06  & {\ 697\(\times\)1359} ~~~& faded ink, lines, degraded background. \\
Persian07  & {\ 637\(\times\)1149} ~~~& faded ink, lines, degraded background, multi-color background.\\
Persian08  & {\ 1617\(\times\)969} ~~~& blur, faded ink, spots, degraded background. \\
  
Persian09  & {\ 1025\(\times\)719} ~~~& faded ink, ink smear, ink noise, alien ink, degraded background.\\
Persian10  & {\ 1649\(\times\)1258} ~~~& lines, blur, multi-color background, visible fibers in paper, bleed-through.\\
\hline

\end{tabular}
\end{center}
\end{table*}
\setlength{\tabcolsep}{1.4pt}

For each image in the dataset, best value of each measure between all of the methods is considered. A method with best value for a measure takes a score of 1, and other methods takes a fraction of 1 by a comparison with the best value. Since there are six evaluation measures and ten test images, we can compute the score of a method as: 

\begin{equation}
S_{k=1}^{3}=\sum_{i=1}^{10} \sum_{j=1}^{6}(\frac {Best_{i,j}}{value_{k,i,j}}~,~\frac {value_{k,i,j}}{Best_{i,j}})~.
\end{equation}

where, \(k\) denotes the number of participators and \(value\) is the measure value obtained by a method. First fraction is used for those measures in which a lower value indicates to better score, and the second one is used for measures with inverse behavior. Finally, methods with higher scores take higher rank. Table II provides detailed experimental results of the binarization algorithms participated in PHIBC 2012 and some state-of-the-art binarization methods. Between three participated algorithms, the algorithm submitted by \textbf {1-} Su Bolan\(^\dagger\), Tian Shangxuan\(^\dagger\), Lu Shijian\(^\ddagger\) and Tan Chew Lim\(^\dagger\) (\(^\dagger\)School of Computing, National University of Singapore, and \(^\ddagger\)Department of Computer Vision and Image Understanding Institute for Infocomm Research, Singapore) achieved the best performance. Figure 2 shows binarization results of the winner of PHIBC 2012.

\begin{table*}[!t]
\renewcommand{\arraystretch}{1.3}
\caption{Evaluation of the binarization methods participated in the PHIBC 2012}
\label{table_example}
\centering
\begin{tabular}{| c | c | c | c | c | c | c | c |}
\hline
\bfseries Method no. &  \bfseries ~Rank~/~Score~ & ~F-Measure~ & ~pseudo F-Measure~ &  ~PSNR~ & ~DRD~ & ~MPM\((\times 10^{-3})\)~ & ~NRM\((\times 10^{-2})\)~\\
\hline\hline
\bfseries ~1~ & \bfseries ~1 / 51.3792~ & ~88.55~ & ~92.25~ & ~18.28~ & 5.57 & 2.33 & 6.84\\
\hline
\bfseries ~2~ & \bfseries ~3 / 50.2433~ & ~86.79~ & ~86.29~ & ~17.64~ & 6.08 & 2.74 & 5.59\\
\hline
\bfseries ~3~ & \bfseries ~2 / 50.7329~ & ~87.30~ & ~89.50~ & ~17.95~ & 5.87 & 3.79 & 5.42\\
\hline\hline
 Otsu \cite{otsu} & - & 77.75 & 79.98 & 15.42 & 31.11 & 16.50 & 5.69\\
\hline
 Grid based Sauvola \cite{farrahimulti} & - & 85.29 & 87.75 & 17.73 & 9.99 & 6.01 & 4.73\\
\hline
ESBK \cite{adotsu} & - & 84.03 & 86.43 & 17.60 & 14.79 & 7.13 & 5.52\\
\hline
 Su's method \cite{Sulmm} & - & 88.21 & 88.82 & 18.27 & 5.44 & 2.65 & 5.74\\
\hline
 PC \cite{ziaeiaccv} & - & 90.19 & 91.35 & 19.89 & 5.23 & 2.74 & 3.90\\
 \hline
Howe \cite{howe2011} & - & 89.58 & 91.88 & 18.53 & 4.11 & 2.96 & 4.56\\
\hline
\end{tabular}
\end{table*}

\begin{figure}[!ht]
\centering

\begin{tabular}{c}
\fbox{{\includegraphics[height = 3.2cm]{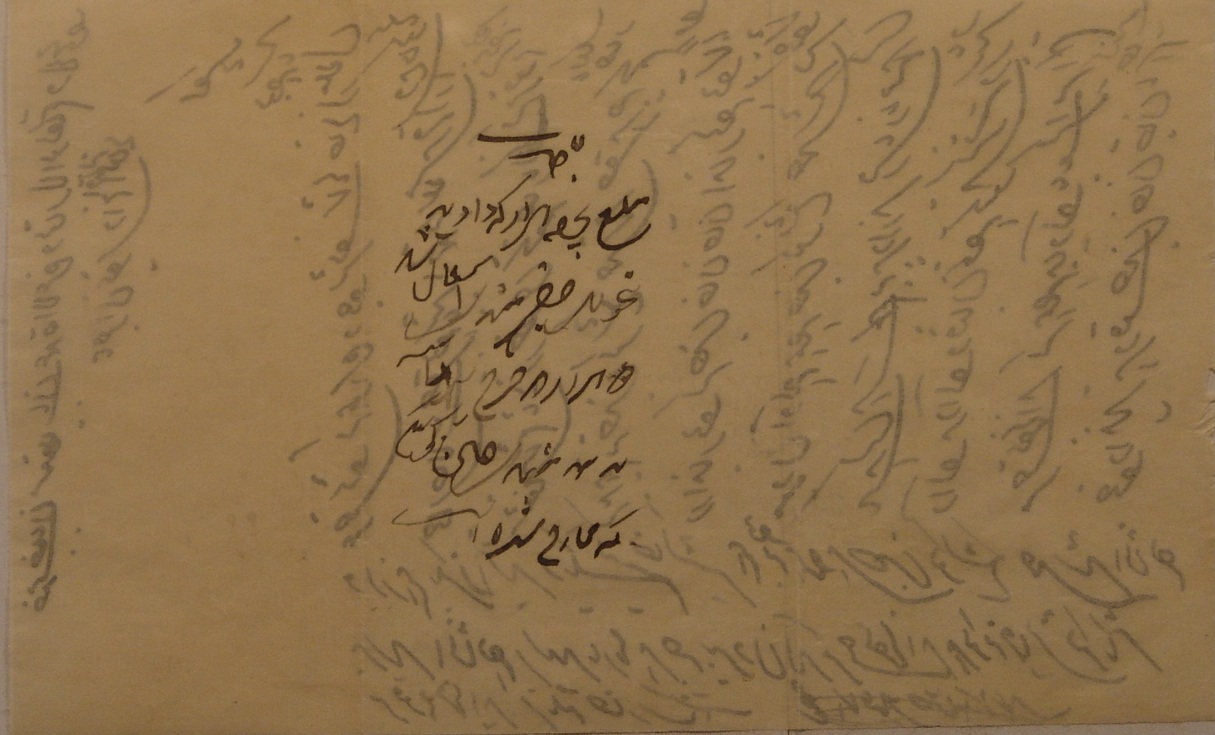}}} \\\\
\fbox{{\includegraphics[height = 3.2cm]{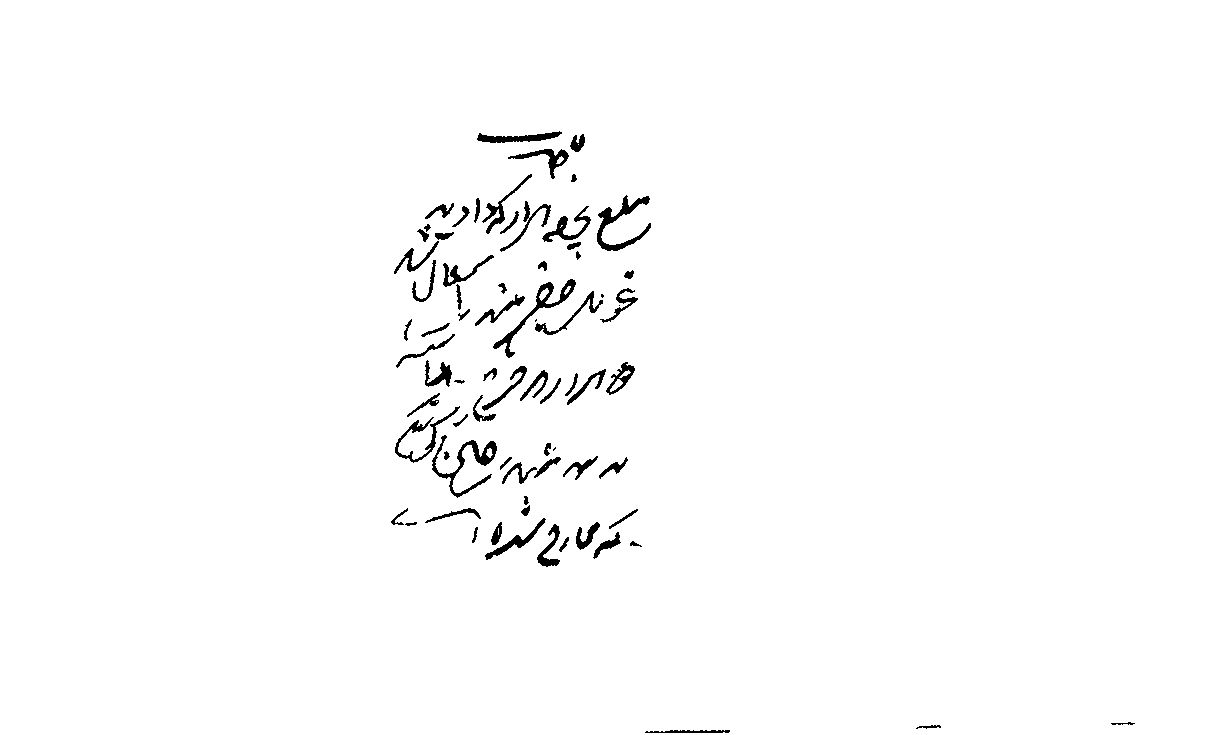}}} 

\\\\
\end{tabular}

\begin{tabular}{cc}
\fbox{{\includegraphics[height = 6cm]{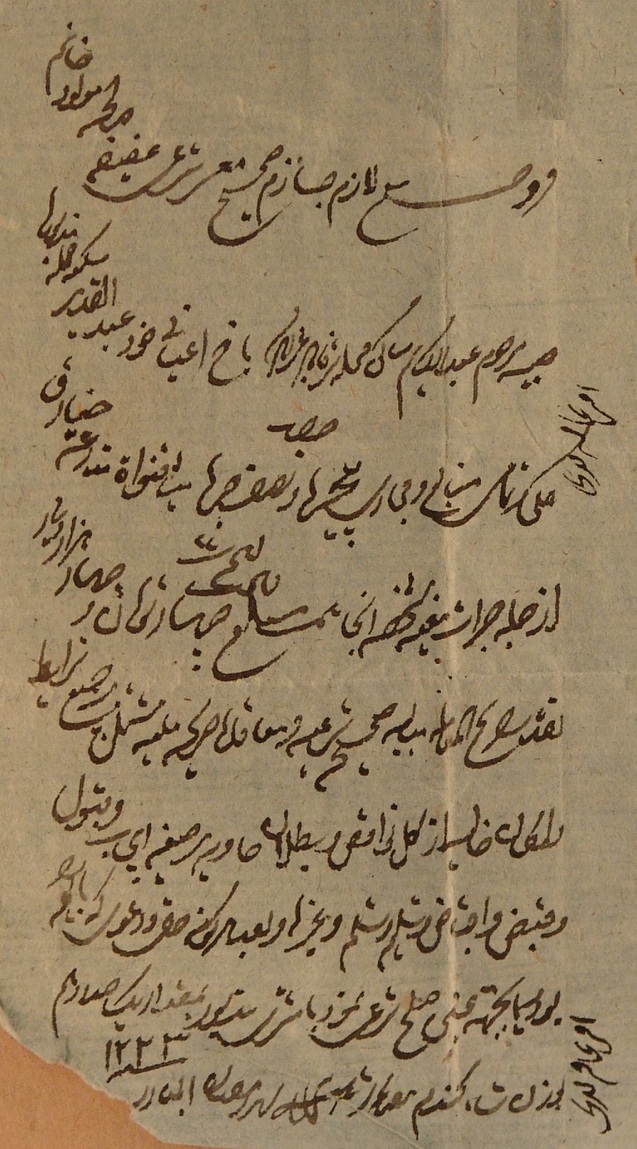}}} &
\fbox{{\includegraphics[height = 6cm]{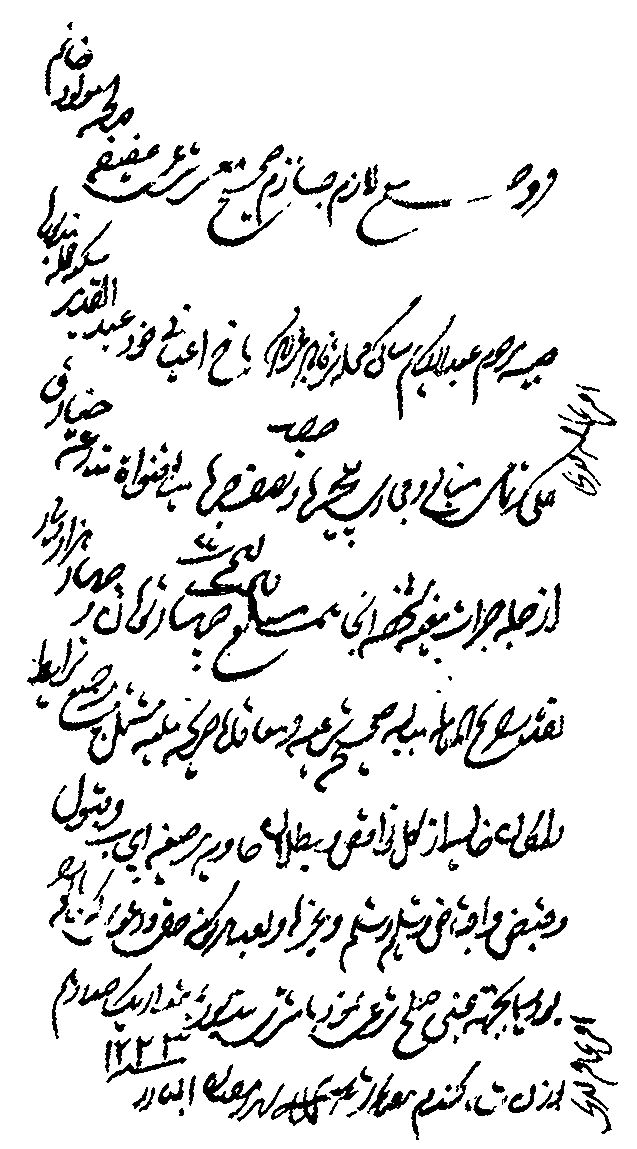}}} 
\end{tabular}

\caption {Sample binarization results from the winner of PHIBC 2012.}
\label{f-outputs}
\end{figure}

\section{Conclusion}

This paper provides a report on the first Persian heritage image binarization competition (PHIBC 2012) which has been organized in conjunction with the first Iranian conference on pattern recognition and image analysis (PRIA 2013). The main objective of this competition is to evaluate the performance of the binarization methods, when applied on the historical Persian document images. The images used in PHIBC 2012 include wide range of degradation types and their associated ground truth are publicly available. Six evaluation measures has been used for comparison between submitted algorithms to PHIBC 2012. Based on the performance of the groups participated in the competition and state-of-the-art binarization methodologies, there is a lot of room for development of higher performance binarization algorithms.


\section*{Acknowledgment}

The organizers of PHIBC 2012 would like to thank "Documents and old manuscripts treasury of Mirza Mohammad Kazemaini (affiliated with Hazrate Emamzadeh Jafar), Yazd, Iran" for providing us the images used in the PHIBC 2012.



%

\bibliographystyle{IEEEtran}
\bibliography{egbib2}

\end{document}